\documentclass[10pt,twocolumn,letterpaper]{article}

\usepackage[pagenumbers]{cvpr} 

\makeatletter
\@namedef{ver@everyshi.sty}{}
\makeatother

\usepackage[accsupp]{axessibility} 
\usepackage{graphicx}
\usepackage{amsmath}
\usepackage{amssymb}
\usepackage{array}
\usepackage{booktabs}
\usepackage{acronym}
\usepackage{subfloat}

\usepackage[pagebackref,breaklinks,colorlinks]{hyperref}

\usepackage[capitalize]{cleveref}
\crefname{section}{Sec.}{Secs.}
\Crefname{section}{Section}{Sections}
\Crefname{table}{Table}{Tables}
\crefname{table}{Tab.}{Tabs.}

\def\cvprPaperID{27} 
\def\confName{CVPR}
\def\confYear{2022}

\usepackage{xspace}
\usepackage{stmaryrd}  
\usepackage{pifont}    
\usepackage[usenames,dvipsnames]{xcolor}
\usepackage{suffix}
\usepackage{todonotes}
\presetkeys{todonotes}{inline}{}  

\newcolumntype{C}[1]{>{\centering\let\newline\\\arraybackslash\hspace{0pt}}m{#1}}
\newcolumntype{L}[1]{>{\raggedright\let\newline\\\arraybackslash\hspace{0pt}}m{#1}}

\newcommand{\mycaption}[2]{\caption[#1]{\textbf{#1.}#2}}

\newcommand{\addfig}[3][!t]{
\begin{figure}[#1]
\centering
\input{#2}
\label{fig:#3}
\end{figure}
}

\WithSuffix\newcommand\addfig*[3][!t]{
\begin{figure*}[#1]
\centering
\input{#2}
\label{fig:#3}
\end{figure*}
}

\newcommand{\addtbl}[3][!t]{
\begin{table}[#1]
\renewcommand{\arraystretch}{1.2}
\centering
\input{#2}
\label{tbl:#3}
\end{table}
}

\WithSuffix\newcommand\addtbl*[3][!t]{
\begin{table*}[#1]
\renewcommand{\arraystretch}{1.2}
\centering
\input{#2}
\label{tbl:#3}
\end{table*}
}

\newcommand{\nbd}{\nobreakdash-} 

\newcommand{\pnorm}[1]{L\ensuremath{_#1}}

\newcommand{\heading}[1]{\noindent \textbf{#1.}}

\newcommand{\fig}[1]{Figure~\ref{fig:#1}}
\newcommand{\tbl}[1]{Table~\ref{tbl:#1}}
\newcommand{\sct}[1]{Section~\ref{sec:#1}}
\newcommand{\eq}[1]{(\ref{eq:#1})}

\newcommand{\myac}[1]{\text{\acs{#1}}}  

\newcommand{\minus}{\scalebox{0.5}[0.75]{\ensuremath{-}}}
\newcommand{\inv}{\minus 1}

\newcommand{\sca}[1]{\ensuremath{#1}}  						   
\newcommand{\mat}[1]{\ensuremath{\textbf{\MakeUppercase{#1}}}} 

\newcommand{\setreal}{\ensuremath{\mathbb{R}}}

\newcommand{\brackets}[3]{\ensuremath{\left#1 #2 \right#3}}
\newcommand{\mypar}[1]{\brackets{(}{#1}{)}}
\newcommand{\mybra}[1]{\brackets{\{}{#1}{\}}}

\newcommand{\easyfunc}[2]{\ensuremath{#1\mypar{#2}}}                 
\newcommand{\func}[3]{\ensuremath{ #1 = \easyfunc{#2}{#3} }}         
\newcommand{\deffunc}[3]{\ensuremath{ \easyfunc{#1}{#2} = #3 }}      

\graphicspath{{Images/}}

\acrodef{msa}[MSA]{Multi-Scale Attention}
\acrodef{sfa}[SFA]{Shared Feature Attention}
\acrodef{mtl}[MTL]{multi-task learning}
\acrodef{rmse}[RMSE]{Root Mean Squared Error}
\acrodef{miou}[m-IoU]{mean-Intersection over Union}
\acrodef{sota}[SotA]{State-of-the-Art}
\acrodef{medusa}[\emph{Medusa}]{Multi-scale Extraction of Dense Universal features via Spatial Attention}
\acrodef{ufl}[UFL]{universal feature learning}

\acrodef{hadamard}[\sca{\odot}]{}
\acrodef{concat}[\sca{\oplus}]{}
\acrodef{scale}[\sca{s}]{}
\acrodef{Scale}[\sca{S}]{}
\acrodef{task}[\sca{t}]{}
\acrodef{Task}[\sca{T}]{}
\acrodef{ch-scale}[\( \sca{C}_{\myac{scale}} \)]{}
\acrodef{att}[\sca{SA}]{}
\acrodef{att-task}[\( \acs{att}_{\myac{scale}}^{\myac{task}} \)]{}
\acrodef{sigmoid}[\sca{\sigma}]{}
\acrodef{conv}[\sca{\phi}]{}
\acrodef{conv-task}[\( \myac{conv}^{\myac{task}} \)]{}
\acrodef{conv-task-scale}[\( \myac{conv}_{\myac{scale}}^{\myac{task}} \)]{}
\acrodef{resblock}[\sca{\psi}]{}
\acrodef{backbone}[\( \mat{B}_{\myac{scale}} \)]{}
\acrodef{fmap}[\mat{F}]{}
\acrodef{fmap-task}[\( \myac{fmap}_{\myac{scale}}^{\myac{task}} \)]{}
\acrodef{fmap-task-ref}[\( \bar{\myac{fmap}}_{\myac{scale}}^{\myac{task}} \)]{}
\acrodef{pred}[\mat{Y}]{}
\acrodef{pred-initial}[\( \myac{pred}_{\myac{scale}}^{\myac{task}} \)]{}
\acrodef{pred-final}[\( \myac{pred}^{\myac{task}} \)]{}
\acrodef{feat-final-scale}[\( \mat{H}_{\myac{scale}}^{\myac{task}} \)]{}
\acrodef{feat-final}[\( \mat{H}^{\myac{task}} \)]{}
\acrodef{multitask}[\sca{m}]{}
\acrodef{baseline}[\sca{b}]{}
\acrodef{mtl-perf}[\( \sca{\Delta}_{\myac{multitask}} \)]{}
\acrodef{perf}[\sca{M}]{}
\acrodef{label-task}[\( \sca{l}^{\myac{task}} \)]{}

\begin{document}

\title{Medusa: Universal Feature Learning via Attentional Multitasking}

\author{Jaime Spencer,\quad Richard Bowden,\quad Simon Hadfield\\
Centre for Vision, Speech and Signal Processing (CVSSP)\\
University of Surrey\\
{\tt\small \{jaime.spencer, r.bowden, s.hadfield\}@surrey.ac.uk}}
\maketitle

\begin{abstract}
Recent approaches to multi-task learning (MTL) have focused on modelling connections between tasks at the decoder level. This leads to a tight coupling between tasks, which need retraining if a new task is inserted or removed. We argue that MTL is a stepping stone towards \emph{universal feature learning} (UFL), which is the ability to learn generic features that can be applied to new tasks without retraining.

We propose \emph{Medusa} to realize this goal, designing task heads with dual attention mechanisms. The shared feature attention masks relevant backbone features for each task, allowing it to learn a generic representation. Meanwhile, a novel Multi-Scale Attention head allows the network to better combine per-task features from different scales when making the final prediction. We show the effectiveness of \emph{Medusa} in UFL (+13.18\% improvement), while maintaining MTL performance and being 25\% more efficient than previous approaches.
\end{abstract}

\section{Introduction}
Classical approaches to computer vision relied on hand-crafted heuristics and features that encapsulated what researchers believed would be useful for a given task. 
With the advent of deep learning, features have become part of the learning process, leading to representations that would have never been developed heuristically.
Unfortunately, most deep learning systems learn features that perform well on only one target task.
Even if pretrained features are used, these require finetuning.
Works that explored generic features~\cite{Detone2018,Dusmanu2019,Spencer2019} have focused on invariance to illumination and viewpoint changes, with the objective of establishing geometric correspondences.
Whilst this is a useful step in many applications, these features are not suitable for a wider range of tasks.


\begin{figure}[!t]
\centering
\input{Figures/intro}
\label{fig:intro}
\end{figure}


Meanwhile, there has been a recent surge in \ac{mtl}, since training a network to solve multiple tasks simultaneously can provide a performance increase over training each task independently~\cite{Maninis2019,Liu2019}. 
Nonetheless, these works have focused on maximizing accuracy, not generality.
At their core, they still try to learn features that perform well on a specific subset of tasks.
It is often difficult or impossible to include new tasks into a previously trained model.
Moreover, modern approaches~\cite{Xu2018,Vandenhende2020d} have such tight connections between tasks that it becomes impossible to evaluate a single task without all other training tasks, as illustrated in \fig{intro}. 
It is difficult to argue that such features are truly generic.

This paper addresses the problem of \ac{ufl}, where a system is capable of learning generic features useful for all tasks.
As discussed, \ac{mtl} is evaluated on the same set of tasks used during training, \ie the training and evaluation tasks are identical.
In contrast, \ac{ufl} aims to generalize \emph{beyond} this training set.
In other words, the training and evaluation tasks are not the same. 
As such, the resulting representations produced by the backbone are referred to as \emph{universal features}.
It is worth noting that, in order to insert a new task into the network, the layers corresponding to the task head still need training. 
However, the focus of \ac{ufl} is on learning backbone feature representations that are left frozen while adding these new task heads.
This results in shorter and more efficient training, as well as avoiding catastrophic forgetting in the shared backbone features. 

The method proposed in this paper, dubbed \acs{medusa}, aims to learn this universal representation.
We design an architecture with completely independent task heads, where the only shared component is the backbone. 
Each task head retains only the specific subset of relevant backbone features via a spatial attention mechanism.
This allows the backbone to learn generic features, while reducing the likelihood of \emph{negative transfer} between tasks.
The model then makes initial predictions at each backbone resolution, which are further combined in a novel \ac{msa} head. 
By feeding back diverse training tasks, we encourage the learned features to encode a wide variety of information across scales.
Furthermore, independent task heads result in an efficient feature extraction process that utilizes significantly less resources but maintains competitive performance, while having a flexible architecture where new task heads can be easily added. 

Our contributions are summarized as:
\begin{enumerate}
\item We highlight the importance of \emph{\acl{ufl}} in contrast to \ac{mtl}.
The main objective behind this is to learn a universal language for computer vision applications. 
This requires a system to learn features that require no additional finetuning to perform well in tasks they were not originally trained for.
In practice, this means that the set of evaluation tasks is different from those used during feature training.

\item We present a novel \acl{msa} task head and show how it can be used to develop an architecture capable of addressing the \ac{ufl} problem.

\item Finally we show that \acs{medusa} can still be applied to traditional \ac{mtl}, where it achieves competitive performance while requiring far fewer resources.
\end{enumerate}

\section{Related Work}
\heading{Multi-task learning}
At its core, \ac{mtl}~\cite{Caruana1997,Ruder2017,Vandenhende2020b} aims to train a single network to accomplish multiple tasks.
Through feature sharing, these models can reduce compute requirements while performing better than expert network counterparts.
Initial approaches consisted of multiple task encoders with additional feature sharing layers. 
The seminal UberNet~\cite{Kokkinos2017} introduced a multi-scale, multi-head network capable of performing a large number of tasks simultaneously.
Cross-stitch networks~\cite{Misra2016} introduced soft feature sharing by learning linear combinations of multiple task features.
In practice, this requires first training each task separately and then finetuning their features.
Sluice networks~\cite{Ruder2019} extended this idea by incorporating subspace and skip-connection sharing. 
Meanwhile, NDDR-CNNs~\cite{Gao2019} replaced the linear combination of features with a dimensionality reduction mechanism.

Kokkinos~\etal~\cite{Kokkinos2017} and Zhao~\etal~\cite{Zhao2018} showed that feature sharing in unreleated tasks results in a degradation in performance for both tasks, known as \emph{negative transfer}.
To account for this, MTAN~\cite{Liu2019} used convolutional attention to build task specific decoders from a shared backbone.  
Other methods learn where to branch from the backbone and what layers to share. 
Vandenhende~\etal~\cite{Vandenhende2020c} decide what layers to share based on precomputed task affinity scores~\cite{Dwivedi2019}.
FAFS~\cite{Lu2017} begins with a fully shared model, optimizing the separation between dissimilar tasks while minimizing model complexity.
BMTAS~\cite{Bruggemann2020} and LTB~\cite{Guo2020} instead use the Gumbel softmax to represent branching points in a tree structure. 

More recent approaches introduce additional refinement steps prior to making the final prediction. 
PAD-Net~\cite{Xu2018} was the first of these networks, using simple task heads to make intermediate predictions.
Each possible pair of tasks were then connected via spatial attention, from which a final prediction was made.
MTI-Net~\cite{Vandenhende2020d} extended this approach to multiple scales, incorporating feature propagation modules between them.
PAP-Net~\cite{Zhang2019a} instead learned per-task pixel affinity matrices, estimating the pixel-wise correlation between each combination of tasks.
Zhou~\etal~\cite{Zhou2020} additionally incorporated inter-task patterns.
Due to the connections between all possible tasks, these approaches suffer from a quadratic growth of network parameters, leading to intractable compute requirements.


\begin{figure}[!t]
\centering
\input{*}
\label{fig:Figures/overview}
\end{figure}
{overview}

\heading{Transfer learning}
A topic closely related to \ac{ufl} is transfer learning~\cite{Torrey2009,Pan2010,Tan2018,Zhuang2021}.
However, these works typically focus on solving domain shift at the input level, performing the same task with a different input modality.
In other cases, the target is a closely related task, \eg~classification on a different set of labels.
More closely related to \acs{medusa} are feature- and network-based techniques for transfer learning.
Feature-based approaches aim to transform the source feature representations into new representations for the target domain. 
This includes approaches such as feature augmentation~\cite{Daume2007,Iii2010,Duan2012}, mapping~\cite{JialinPan2008,Pan2011,Long2013}, clustering~\cite{Dai2007,Dai2008} and sharing~\cite{Lee2007,Davis2009,Gong2012}.
Meanwhile, network-based techniques have instead focused on parameter sharing. 
Some notable examples include matrix factorization~\cite{Zhuang2011,Zhuang2014} and parameter reuse from a network pretrained on a source domain~\cite{Oquab2014,Thompson2019}.
However, the focus lies mainly on the performance after finetuning on a specific new target task, rather than the overall performance on a wide range of tasks, which is the focus of \ac{mtl} and \ac{ufl}.

More recently,~\cite{Zamir2018,Dwivedi2019} were proposed to learn and model the relationship between tasks.
However, these approaches do not truly solve the \acs{ufl} problem since it is still necessary to train a separate network on each task.
Instead they follow a brute force approach to find the best possible source to transfer for a given target task. 
In contrast, \acs{medusa} learns a single representation which can generalize well across future target tasks.

\section{Methodology}
The aim of this work is to introduce an architecture capable of learning universal features that perform well in multiple different tasks.
As shown in \fig{overview}, \acs{medusa} consists of two main components: a shared backbone and individual task heads.
One key design feature is that each task head is independent from the rest. 
This allows us to add new task heads \emph{a posteriori}, which can be trained in conjunction or separately from the existing tasks.

\subsection{Shared Feature Attention}
The only part of the architecture common to all tasks is the shared backbone.
The backbone produces the shared features \( \acs{backbone} \in  \setreal^{\scriptsize \acs{ch-scale}} \)
at multiple scales \acs{Scale}, where \acs{ch-scale} is the number of channels per-scale.
Our goal is to learn universal features useful in a wide range of tasks, which may not be known at training time.
In order to let the backbone learn a broad range of features whilst allowing tasks to pick their own specific subsets, we introduce spatial attention between the backbone and each task head.
We define the process of applying spatial attention \acs{att} to a generic feature map \acs{fmap} as 
\begin{equation}  \label{eq:spatial_att}
\deffunc
	{ \acs{att}  }
	{ \acs{fmap} }
	{
	  \easyfunc{ \acs{sigmoid} }{ \easyfunc{ \acs{conv}_1 }{ \acs{fmap} } }
	  \acs{hadamard}
	  \easyfunc{ \acs{conv}_2 }{ \acs{fmap} }
	}
,
\end{equation}
where \acs{sigmoid} is the sigmoid operation, \acs{hadamard} the Hadamard product and \acs{conv} a convolution operation followed by batch normalization and a ReLU activation. 
Note that the concept of spatial attention is also known as the GLU activation~\cite{Dauphin2017} and has previously been used in \ac{mtl}~\cite{Liu2019,Xu2018,Vandenhende2020d}.
In \acs{medusa}, the convolution weights for each scale and task are independent from each other.
Therefore, 
\(
\func
	{ \acs{fmap-task} }
	{ \acs{att-task} }
	{ \acs{backbone} }
\)
represents the initial task features for scale \acs{scale} and task \acs{task}.

The shared backbone can now learn a generic feature representation that suits a much wider range of tasks.
Through the per-channel spatial attention 
\(
\easyfunc
	{ \acs{sigmoid} }
	{ \easyfunc{ \acs{conv}_1 }{ \acs{fmap} }}
\), 
each task/scale retains only the specific subset of backbone features relevant to it. 
This alleviates the possibility of \emph{negative transfer}, where sharing features between unrelated tasks can degrade the performance of both tasks.
Whilst previous approaches also make use of spatial attention, they place a larger focus on modeling the connections between each pair of tasks. 
By creating an information bottleneck, \acs{medusa} places more importance on learning features common to all tasks that therefore provide better transfer capabilities. 
Additionally, our multi-scale approach provides the subsequent task features with a wide variety of information which, combined with the proposed \ac{msa} head, helps to provide optimal features for the final prediction. 

\subsection{Multi-Scale Task Predictions}
Rather than building a per task sequential decoder such as~\cite{Liu2019}, we build parallel task heads by using each scale of backbone features to make an initial prediction for each task. 
This results in \acs{Scale} predictions per task, used as additional supervision during training.
The initial task features \acs{fmap-task} are refined through
\begin{equation}
\func
	{ \acs{fmap-task-ref}  }
	{ \acs{resblock}_2 }
	{ \easyfunc{ \acs{resblock}_1 }{ \acs{fmap-task} } }
,
\end{equation}
where \acs{fmap-task-ref} are the refined task features and 
\(
\deffunc
	{ \acs{resblock} }
	{ \acs{fmap} }
	{ \easyfunc{ \acs{conv} }{ \acs{fmap} } + \acs{fmap} }
\)
is a residual convolutional block.
The initial predictions are given by 
\( 
\func
	{ \acs{pred-initial} }
	{ \acs{conv-task-scale} }
	{ \acs{fmap-task-ref} }
\), 
where \acs{conv-task-scale} is the convolution mapping from \acs{ch-scale} channels provided by the backbone to those required by the task.
These predictions are used as intermediate supervision exclusively during training, while the refined task features are combined in the \ac{msa} heads.

\subsection{Multi-Scale Attention Task Heads}
The final step is combining task features from multiple scales to make the final prediction for each task. 
In the na\"ive case, one could simply upsample all task features to the same resolution, concatenate channel-wise and process them together~\cite{Vandenhende2020d}.
We refer to this task head as HRHead in the results further on. 
This assumes that the predictions from each scale are equally valid and important. 
However, due to the varying resolution and subsequent receptive field of each scale, this is not typically the case. 
In practice, higher resolution predictions can help to provide more accurate and sharp edges for some tasks. 
On the other hand, lower resolutions with more channels provide more descriptive features with a larger receptive field, making predictions more consistent on a global scale. 

We capture this information by introducing a novel \acl{msa} task head. 
Given the processed task features \acs{fmap-task-ref} the network is able to select the important information from each scale using the spatial attention \acs{att} previously defined in \eq{spatial_att}. 
This results in
\begin{equation}
\func
	{ \acs{feat-final-scale}  }
	{ \acs{att-task} }
	{ \acs{fmap-task-ref} }
,  
\end{equation}
\vspace{-0.4cm}
\begin{equation}
\acs{feat-final}
= 
\mat{H}_{\scriptsize 0}^{\scriptsize \acs{task}} 
\acs{concat} 
\mat{H}_{\scriptsize 1}^{\scriptsize \acs{task}} 
\acs{concat} \ldots \acs{concat} 
\mat{H}_{\scriptsize \acs{Scale}}^{\scriptsize \acs{task}} 
, 
\end{equation}
where \acs{concat} represents channel-wise concatenation of the attended per task per scale features \acs{feat-final-scale}.
Note that the spatial attention weights are independent from those previously used to extract \acs{fmap-task}.
The final per task features \acs{feat-final} are used to obtain the final predictions as 
\( 
\func
	{ \acs{pred-final} }
	{ \acs{conv-task} }
	{ \acs{feat-final} }
\), 
where \acs{conv-task} maps the final number of channels \( \sum \acs{ch-scale} \) to the required task channels.

Thanks to the design of the system it becomes trivial to attach new task heads to the shared backbone.
These task heads are able to choose relevant features from the shared backbone and adapt the multiple scales to the needs of new tasks. 
Furthermore, since the task heads are independent, the number of parameters increases only linearly with the number of tasks.
Because these heads are lightweight, the resulting system is highly efficient.
This is contrary to approaches such as~\cite{Vandenhende2020d,Xu2018}, where each task requires connections to every other task, resulting in a quadratic parameter-complexity with regards to the number of tasks.

\section{Results}
\heading{Dataset}
We use the NYUD-v2 dataset~\cite{Silberman2012}, containing labels for depth estimation, semantic segmentation, edge detection and surface normal estimation. 
Following existing benchmarks~\cite{Maninis2019,Vandenhende2020d}, we focus on evaluating depth and semantic segmentation, leaving edges and surface normals as auxiliary tasks for use during training.
Depth is evaluated through the \ac{rmse}, while semantic segmentation uses the \ac{miou}.

\heading{Implementation details}
We use HRNet-18~\cite{Sun2019} pretrained on ImageNet~\cite{Deng2010} as the backbone, due to its suitability for dense prediction tasks.
This produces features at downsampling scales of \mybra{4, 8, 16, 32} with \mybra{18, 36, 72, 144} channels, respectively.
We use the Adam optimizer, with a base LR=1e-4 and a polynomial decay~\cite{Chen2018b}.
Experimentally, we found that training the shared backbone with a lower learning rate than the heads (typically LR*0.1) produced better results. 
Models are trained for 100 epochs.
Regarding the losses, we use the \pnorm{1} loss for depth and surface normal estimation, cross-entropy for semantic segmentation and a binary cross-entropy (with positive weighting of 0.95) for edge detection. 

\subsection{Multi-task Evaluation} \label{sec:mtl}


\begin{table}[!t]
\renewcommand{\arraystretch}{1.2}
\centering
\input{*}
\label{tbl:Tables/multi_task}
\end{table}
{multi_task}


\begin{table}[!t]
\renewcommand{\arraystretch}{1.2}
\centering
\input{*}
\label{tbl:Tables/ablation}
\end{table}
{ablation}


\begin{figure}[!t]
\centering
\input{*}
\label{fig:Figures/qualitative}
\end{figure}
{qualitative}

\heading{Performance}
We first evaluate \acs{medusa}'s performance in a traditional \ac{mtl} setting, following the procedure in~\cite{Maninis2019}.
As mentioned, the main tasks evaluated are depth estimation and semantic segmentation. 
However, during training we make additional use of \textbf{E}dge detection and surface \textbf{N}ormal estimation to show the network a varied set of tasks. 
Following~\cite{Maninis2019}, we define multi-task learning performance as 
\begin{equation} \label{eq:mtl}
\acs{mtl-perf}
=
\frac{1}{\acs{Task}} 
\sum_{\scriptsize \acs{task}=0}^{\scriptsize \acs{Task} \minus 1} 
\mypar{\inv}^{\scriptsize \acs{label-task}} 
\frac
	{ 
	  \acs{perf}_{\scriptsize \acs{multitask}}^{\scriptsize \acs{task}} 
	  \minus 
	  \acs{perf}_{\scriptsize \acs{baseline}}^{\scriptsize \acs{task}} 
	}
	{ \acs{perf}_{\scriptsize \acs{baseline}}^{\scriptsize \acs{task}} }
,
\end{equation}
where $ \acs{perf}^{\scriptsize \acs{task}}_{\scriptsize \mybra{\acs{multitask}, \acs{baseline}}} $ is the per-task performance of the \emph{\textbf{m}}ultitask or \emph{\textbf{b}}aseline network and \acs{label-task} indicates if a lower value means a better performance for the given task.
As such, \acs{mtl-perf} represents the average increase (or drop) in performance for each task, relative to the single task baseline.

We obtain single task baselines (ST) for each backbone by training expert networks on each task separately, resulting in two completely separate models.
ResNet models use Deeplab-v3+~ASPP~\cite{Chen2018c} task heads, while HRNet-18 uses the na\"ive multi-scale task head, upsampling all scales and concatenating channel-wise (HRHead).
Meanwhile, the multi-task baselines (MT) use a joint backbone with separate task heads. 
The baselines were obtained by retraining the code provided by the authors of~\cite{Maninis2019,Vandenhende2020d}.
In order to make results more comparable, we also create and train a version of MTAN adapted to make use of the \mbox{HRNet} backbone. 
However, since MTAN builds a per task decoder, rather than making initial predictions at multiple scales, it still requires use of the DeepLap-v3+ head.  

The results can be found in \tbl{multi_task}, where the column (N+E) indicates the presence of the auxiliary edges and surface normals tasks.
It is interesting to note that some MT baselines and methods~\cite{Gao2019,Xu2018} actually lead to a degradation in performance. 
This is likely due to a combination of \emph{negative transfer} and task loss balancing during training.
Meanwhile, despite not modelling task connections in the decoder, \acs{medusa} still shows improvements when incorporating the auxiliary (N+E) tasks.
This demonstrates the ability of \acs{medusa} to learn generic features that complement all tasks, sharing only the useful information. 
To summarize, \acs{medusa} greatly outperforms all baselines with independent task heads~\cite{Liu2019} and is comparable to the current \ac{sota}~\cite{Vandenhende2020d} while using resources in a more efficient manner, as we will now discuss.

\heading{Ablation}
We perform an ablation study to understand the importance of \acs{medusa}'s components, primarily focused on the uses of spatial attention.
In the case of the \ac{sfa}, we replace the spatial attention connecting the shared backbone to each task head with a convolutional block with BatchNorm and ReLU. 
On the other hand, we compare the proposed \acs{msa} head to the default HRNet, which does not contain spatial attention. 

\tbl{ablation} shows that both attention components result in large benefits. 
Incorporating the \ac{sfa} results in a consistent relative improvement across the different techniques of 8.94\% and 16.81\%.
Meanwhile, the \ac{msa} head leads to even larger performance gains---from 17.88\%~to~180.60\%.
Most notably, incorporating the \ac{msa} head into the MT baseline results in an improvement over the ST baseline.
Even across all \acs{medusa} variants, this novel task head leads to a consistent increase in accuracy. 
Similarly, incorporating the \ac{sfa} between the backbone and task heads improves performance regardless of the task head used.
Overall, the dual attention mechanisms used in \acs{medusa} lead to a relative improvement of 37.7\% over the plain convolutional baseline. 
Once again, we believe that this is due to spatial attention providing an effective, yet efficient mechanism for routing information between different stages in the network.
This allows it to easily decide what information should or should not be shared across either tasks and scales. 

\heading{Visualizations}
\fig{qualitative} shows qualitative results based on the network predictions. 
As expected, the MT baseline shows the worst results.
MTI-Net shows more spurious class predictions, especially in cluttered environments, as seen in the second image in \fig{qualitative}.
Meanwhile, we find \acs{medusa} to be more globally coherent, while still having well defined edges between classes and in depth discontinuities. 
This is due to the proposed \ac{msa} head, which can effectively combine the best features from each scale. 


\begin{figure}[!t]
\centering
\input{*}
\label{fig:Figures/resources}
\end{figure}
{resources}

\newpage
\heading{Resources}
\fig{resources} shows how different approaches scale \wrt the number of tasks.
The most resource efficient approach is the MT baseline. 
However, since it only uses basic task heads without intermediate predictions or attention, its performance is lacklustre. 
Other approaches with independent task heads (ST, MTAN) are relatively efficient, since the increase in task head parameters is linear, but their performance (see \tbl{multi_task}) is not on par with \acs{medusa}.
MTI-Net is the only approach with results comparable to \acs{medusa}, but it does not scale well to increasing numbers of tasks.
After only three tasks, MTI-Net requires more parameters than the ST baseline, which trains a completely separate network for each task. 
This gap only increases, due to the quadratic parameter-complexity introduced by the connections between all possible pairs of tasks.

\subsection{Universal Feature Learning}


\begin{table}[!t]
\renewcommand{\arraystretch}{1.2}
\centering
\input{*}
\label{tbl:Tables/ufl}
\end{table}
{ufl}

The following experiment evaluates \acs{medusa} in the highlighted \ac{ufl} task. 
The objective is to learn generic shared features that can be adapted to new, unseen tasks on unseen datasets without additional finetuning at the backbone level. 
This is contrary to \ac{mtl}, where the objective is to learn features that perform well in the specific set of training tasks without generalization to other tasks.
It is also contrary to traditional transfer learning, where the objective is 
instead to solve the domain shift between different modalities of a single task.

We show the ability of \acs{medusa} features to transfer to new tasks on new datasets through the PASCAL-Context~\cite{Chen2014} dataset, containing semantic segmentation, human part segmentation and edge detection. 
There are also pseudo-ground truth labels for surface normals and saliency~\cite{Maninis2019} obtained from \acs{sota} models~\cite{Bansal2017,Chen2018c}.
Since three of the tasks are common to NYUD-v2 we evaluate on the two unique ones: human part segmentation and saliency estimation. 
To carry out this evaluation we use the previous models trained on NYUD{\nbd}v2 in \sct{mtl} with the auxiliary (N+E) tasks and check their transfer capability to the new target tasks in the PASCAL{\nbd}Context dataset.
This is done by freezing the shared feature backbone network and adding a new task head corresponding to either saliency estimation or human part segmentation.
This can be seen as a form of continual learning. 
Since the shared backbone and previous task heads are frozen, we ensure that the network does not forget existing information. 
Instead, we expand its knowledge by learning a new task. 

\tbl{ufl} shows the results from this experiment, including the previous \ac{mtl} results on NYUD{\nbd}v2 for comparison.
This highlights the main difference between \ac{ufl} and \ac{mtl}, where \ac{mtl} only performs well in the original training tasks.
This is only exacerbated by the na\"ive multi-task implementation, resulting in a large amount of negative transfer between tasks. 
Meanwhile, \acs{medusa} provides the best transfer capabilities. 
It is worth noting the large improvement over ImageNet pretrained features from the single task baseline (ST), which are trained on orders of magnitude more data than the remaining \ac{mtl} methods.
However, since they are trained exclusively for global image classification, the learnt representations do not transfer well to complex dense tasks. 
Meanwhile, even through  \ac{mtl} performance is almost equal to MTI-Net (9.24\% \vs 9.48\%), the features learnt by \acs{medusa} generalize to a broader range of tasks (13.18\% \vs 10.76\%). 
This is due to \acs{medusa}'s design, which places a larger focus on the shared feature representation, which is therefore able to learn a more effective feature representation.

\section{Conclusions \& Future Work}
In this paper we have highlighted the importance of \emph{\acl{ufl}} \vs \acl{mtl}, requiring a feature learning system to perform well over a large variety of tasks without additional finetuning. 
This is in contrast to most current \ac{mtl} approaches, which focus learning features specific to a given set of training tasks.

To this end we proposed \acs{medusa}, capable of training on multiple tasks simultaneously, while allowing new task heads to be attached and trained jointly or separately. 
Furthermore, thanks to the novel \ac{msa} head, we are capable of doing this in a very efficient manner.
This helps to provide comparable results whilst using less resources than previous approaches. 
We additionally demonstrated the generality of the features leant by \acs{medusa} in the \ac{ufl} task on unseen tasks and datasets, and showed its ability to outperform \ac{sota} features from both ImageNet and other \ac{mtl} networks. 

Whilst \acs{medusa} has shown its effectiveness in both \ac{mtl} and \ac{ufl}, it is not without limitations and challenges to address in future work. 
For instance, the data used during training is currently required to have labels for all target tasks. 
In practice, these labels can be challenging to obtain, especially as the number of tasks and images grows. 
\acs{medusa}'s performance is also dependent on the tasks used during training.
If we wish to transfer to a task that is completely unrelated to the training tasks, it is likely that the features will not overlap. 
Both of these issues could potentially be addressed by making the training process more flexible, without requiring each item to have labels for all tasks or by training with multiple unrelated datasets.

{\small
\bibliographystyle{ieee_fullname}
\bibliography{Medusa_CVPR2022}
}

\end{document}